\documentclass{article}

\usepackage{arxiv}

\usepackage{blindtext}
\usepackage{comment}
\usepackage[T1]{fontenc}
\usepackage[hidelinks]{hyperref}
\usepackage{url}
\usepackage{booktabs}
\usepackage{microtype}
\usepackage{graphicx}
\usepackage{doi}
\usepackage{tabularx}
\usepackage{authblk} 
\usepackage{setspace}
\usepackage[table]{xcolor}
\usepackage{multirow}
\usepackage{amsfonts}
\usepackage{nicefrac}
\usepackage{tikz}
\usepackage{adjustbox}
\usepackage{float}
\usepackage{amsmath}
\usepackage{cleveref}

\definecolor{greenishyellow}{rgb}{0.68, 1.0, 0.18}
\definecolor{lightorange}{rgb}{1.0, 0.6, 0.0}
\definecolor{orangered}{rgb}{1.0, 0.27, 0.0}

\newcommand{\papertitle}{Unsupervised Learning Approaches for Identifying ICU Patient Subgroups: Do Results Generalise?}

\usepackage{fancyhdr}
\usepackage[square, numbers]{natbib}
\setcitestyle{citesep={,}}

\title{Unsupervised Learning Approaches for Identifying\\ ICU Patient Subgroups: Do Results Generalise?} 

\author[1]{Harry Mayne}
\author[1, 2]{Guy Parsons}
\author[1]{Adam Mahdi}
\affil[1]{Oxford Internet Institute, University of Oxford}
\affil[2]{NIHR Academic Clinical Fellow at University of Oxford and Thames Valley Deanery}

\date{}
\usepackage{subfiles} 

\begin{document}

\maketitle

\begin{abstract}
The use of unsupervised learning to identify patient subgroups has emerged as a potentially promising direction to improve the efficiency of Intensive Care Units (ICUs). By identifying subgroups of patients with similar levels of medical resource need, ICUs could be restructured into a collection of smaller subunits, each catering to a specific group. However, it is unclear whether common patient subgroups exist across different ICUs, which would determine whether ICU restructuring could be operationalised in a standardised manner. In this paper, we tested the hypothesis that common ICU patient subgroups exist by examining whether the results from one existing study generalise to a different dataset. We extracted 16 features representing medical resource need and used consensus clustering to derive patient subgroups, replicating the previous study. We found limited similarities between our results and those of the previous study, providing evidence against the hypothesis. Our findings imply that there is significant variation between ICUs; thus, a standardised restructuring approach is unlikely to be appropriate. Instead, potential efficiency gains might be greater when the number and nature of the subunits are tailored to each ICU individually. 
\end{abstract} 
\vspace{5mm}

% %%%%%%%%%%%%%%%%%%%%%%%%%%%%%%%%%%%%%%%%%%%%%%%%%%%%%%%%%%%%%%%%%
\section{Introduction}

Intensive Care Units (ICUs) face growing demand as a result of the growth of an older, more medically comorbid population \cite{creagh-brown_increasing_2014} and through medical and surgical advances placing greater strain on critical care resources \cite{de_lange_icu_2020}. This greater strain on resources can compromise the effectiveness of the care provided \cite{lapichino_volume_2004}. Whilst investment in greater resources is undoubtedly required to support future demand \cite{laake_impact_2010}, intensive care provision is particularly expensive, so finding ways to use existing resources more efficiently should also be considered. 

Intensive care patients comprise a highly heterogeneous population with different illness severities and clinical trajectories \cite{castela_forte_identifying_2021, werner_identification_2023}. This level of heterogeneity can make the efficient provision of intensive care challenging, as clinicians are required to provide specialist care across a generalist scope, and careful judgement is needed to make the most judicious use of resources. One proposal to improve the efficiency of care provision is to use unsupervised learning to cluster together patients with similar levels of medical resource need, which then facilitates the physical restructuring of ICUs into a collection of subunits, each caring for a specific cluster of patients with more homogeneous medical resource need \cite{vranas_identifying_2017, bohmer_care_2008}. This idea is sometimes referred to in the literature as creating \textit{care platforms} \cite{bohmer_care_2008}. Such restructuring could theoretically allow resources to be optimally reallocated so that each subunit provides a level of care that matches their cluster's level of need. This could avoid large under- or over-provision of resources, and more patients could potentially be cared for with a given supply of resources. It is worth noting that this restructuring approach is  different from subspeciality ICUs, such as a Cardiothoracic or Neurosurgical ICUs, which group patients by their diagnosis type rather than by their level of need.

Existing work has progressed this idea. Bohmer and Lawrence \cite{bohmer_care_2008} first proposed improving ICU efficiency by separating patients based on medical need. Vranas et al. \cite{vranas_identifying_2017} suggested the use of unsupervised learning and showed that consensus clustering, a robust ensemble clustering method, could be used to derive meaningful patient subgroups. Subsequent studies further explored the use of unsupervised learning to produce patient subgroups, albeit not exclusively with the aim of ICU restructuring \cite{castela_forte_identifying_2021, werner_identification_2023, hyun_exploration_2020, merkelbach_novel_2023, shi_identifying_2023, fuest2023clustering}. Notably, Castela Forte et al. \cite{castela_forte_identifying_2021} compared the performance of four clustering methods and explored how Shapley values could be used to assess feature importance in the clustering. Additionally, Merkelbach et al. \cite{merkelbach_novel_2023} trained a gated recurrent unit autoencoder to represent irregular and sparse ICU data in a low-dimensional feature space prior to clustering, allowing their clusters to capture a more comprehensive view of ICU stays.

Whilst many studies successfully demonstrate that unsupervised learning can derive meaningful patient subgroups, it remains unclear how well results from individual studies generalise to other datasets \cite{fuest2023clustering, castela_forte_use_2019, de2024deep}. Understanding the generalisability of patient clustering is important for determining how ICU restructuring might be best operationalised. If the results from individual studies generalise and similar patient subgroups can be consistently identified in new datasets, then it might be possible to restructure all ICUs into a standardised set of subunits. If this hypothesis holds, it would offer significant practical advantages, since ICU restructuring could be easily applied to additional ICUs with low marginal costs. On the other hand, if clustering results do not generalise, then a standardised approach may be inappropriate. In this case, it might be better to tailor the number and nature of the subunits to each individual ICU.

Generalisability has recently been explored by de Kok et al. \cite{de2024deep}, who assessed whether the clusters identified by Castela Forte et al. \cite{castela_forte_identifying_2021} could be identified in a different ICU. They trained a deep embedded clustering model on one dataset, then applied the model, without retraining, to a new dataset. This showed that extrapolating clusters across datasets can reveal similar clusters. However, extrapolation is a narrow test for generalisability, since it does not consider whether the new clusters are intrinsic to the new dataset. Our definition of generalisability is stricter than in \cite{de2024deep} and more aligned with reproducibility. In this study, we carry out an experiment to test the hypothesis that common patient subgroups exist across different ICUs. Specifically, we test whether the results from Vranas et al. \cite{vranas_identifying_2017} generalise when the clustering methodology is applied to data from a different ICU. Whilst Vranas et al. \cite{vranas_identifying_2017} derived patient subgroups using data from 21 Kaiser Permanente Northern California hospitals in California, USA, we use the MIMIC-IV dataset, collected at the Beth Israel Deaconess Medical Center in Massachusetts, USA. \cite{johnson_mimic-iv_2023}. First, we replicate the original clustering methodology as closely as possible to ensure that the data is the only source of variation, then we compare our derived clusters with those found in the original study to assess generalisability. Our findings contribute to the understanding of ICU patient clustering and offer insights into the operationalisation of ICU restructuring.

% %%%%%%%%%%%%%%%%%%%%%%%%%%%%%%%%%%%%%%%%%%%%%%%%%%%%%%%%%%%%%%%%%
\section{Methods}

\subsection{Study design}
Our study wished to test the hypothesis that common patient subgroups exist across different ICUs. If this hypothesis were true, it would imply that clustering could identify similar patient subgroups in two different datasets if applied in a consistent way. To test this implication, we replicated the methodology from Vranas et al. \cite{vranas_identifying_2017} in a different dataset, and then assessed the similarities between the clustering results. To ensure that this was a valid test of generalisability, we required the data to be the only source of variation. We therefore took considerable time and care to, first, identify and recreate the same clustering features as in \cite{vranas_identifying_2017}, second, exactly replicate their clustering methodology, and third, interpret our clustering results in the same way. If the method failed to identify the same patient subgroups in the new dataset, it would offer evidence that the results do not generalise.

\subsection{Data sources and inclusion criteria}
We used the MIMIC-IV (version 2.2) dataset \cite{johnson_mimic-iv_2023}, an extensive electronic health record detailing patients and hospitalisations at the Beth Israel Deaconess Medical Center in Boston, Massachusetts. Patients younger than 18 at the time of hospital admission, obstetric patients, and patients known to require enhanced protection were excluded during the creation of the dataset. Details about informed consent and data availability are described in \cite{johnson_mimic-iv_2023}.

\subsection{Feature selection and preprocessing}
\begin{figure}[t!]
    \centering
    \includegraphics[width=\textwidth]{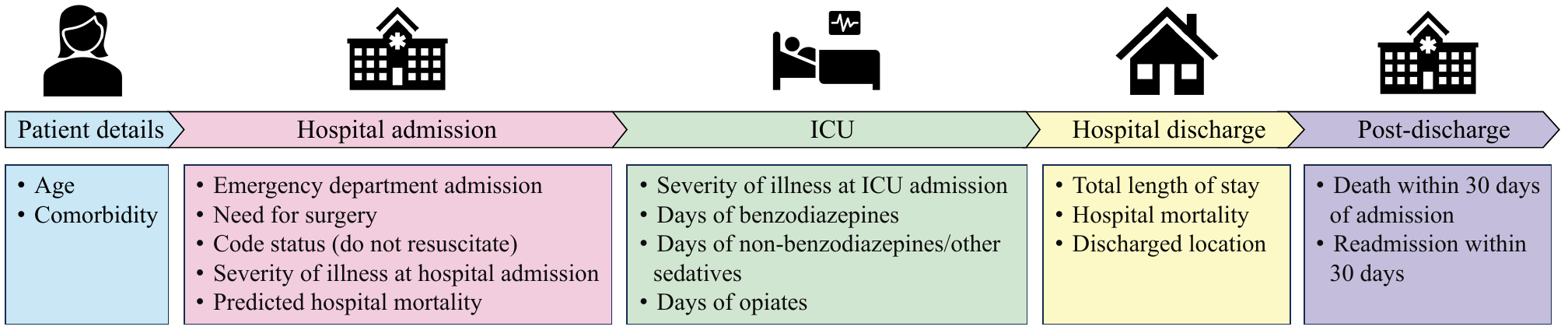}
    \vspace{-5mm}
    \caption[]{\color{black}{\bf Features used to derive the clusters.} The clustering features can be separated into five domains: patient details, hospital admission, ICU, hospital discharge and post-discharge. The clustering features span the duration of patients' hospitalisation to ensure that the resulting clusters represent medical need throughout ICU rather than at a specific point in time. Notably, this means that direct patient triage at ICU admission would not be possible and this issue is discussed further in Section \ref{sec:limitations_future_research}.}
    \label{fig:timeline}
\end{figure}

To create the clustering dataset, we extracted 16 features across five domains: patient details, hospital admission, ICU, hospital discharge and post-discharge (Figure \ref{fig:timeline}). The features spanned the duration of patients' hospitalisations, since we wanted patients to be grouped based on their levels of medical need across their entire ICU stays rather than at a specific point in time. A notable restriction of this approach is that patients cannot be directly triaged to subunits at ICU admission because some features can only be known retrospectively. However, we propose that identifying comprehensive patient subgroups first, and then designing triage methods subsequently, has the potential to ultimately lead to better subgroups than if subgroups are based solely on features available at ICU admission. Approaches to patient triage are discussed further in Section \ref{sec:limitations_future_research}.

In most cases, MIMIC-IV contained near-identical features to those in \cite{vranas_identifying_2017}. However, in some cases, we used more granular data to build analogous features. Where permissible, missing data were imputed with default values, such as `full code' for code status. Additionally, the data were passed through a series of filters to detect outliers and contradictory information. Detailed information about feature creation and preprocessing can be found in Sections \ref{App:feature_matching} and \ref{App:preprocessing} of the Appendix, respectively.

\subsection{Consensus clustering of ICU patients}
We clustered the patients using consensus clustering, leveraging its advantages over traditional single-iteration techniques \cite{monti_consensus_2003}. Consensus clustering is an ensemble clustering method, where a traditional clustering algorithm is repeated many times (\textit{the inner loop}) and the results from the iterations are aggregated to produce a consensus (\textit{the consensus stage}) \cite{strehl2002cluster}. Each iteration uses a slightly perturbed version of the dataset, where only a proportion of the features and examples are selected. Since true patient subgroups should be robust to small changes in the features or examples, consensus clustering can identify meaningful subgroups. Additionally, comparing the stability of different clustering solutions across the iterations offers a good way to select the most appropriate number of clusters.

Our specific implementation is as follows. First, we randomly selected a sample of 5,000 ICU stays from the total cohort, since the algorithm's run time scales approximately quadratically with the number of examples \cite{Manning2008IntroductionTI}. To avoid imbalanced influence, all features, including binary ones, were standardised in the traditional z-score manner. We then employed consensus clustering, using agglomerative hierarchical clustering with average linkage \cite{rr1958statiscal} for both the inner loop and consensus stages. The inner loop included 1,000 iterations, each time sampling 80\% of features and ICU stays. This was used to derive eight different clustering solutions, where we varied the number of clusters $K$, from $2$ to $9$ (henceforth referred to as the \textit{clustering solutions}). 

To select the most appropriate clustering solution, we examined the stability of each solution across the 1,000 iterations \cite{monti_consensus_2003, senbabaoglu_critical_2014, wilkerson_consensusclusterplus_2010}. We considered visualisations of the ordered consensus matrices, the cumulative distribution functions (CDFs) of the consensus indices and the tracking plot to identify cases where the clustering solutions, or parts of the solutions, were likely to be unstable and thus not representative of meaningful patient subgroups. Notably, we did not force our final choice to have the same number of clusters as Vranas et al. \cite{vranas_identifying_2017}, but rather, selected the most appropriate solution for our dataset. This ensured that our results were a true reflection of the subgroups intrinsic to our data, rather than potentially forcing a poorer clustering solution. The clusters and visualisations were generated using the \texttt{R} library \texttt{ConsensusClusterPlus} \cite{wilkerson_consensusclusterplus_2010}.

\subsection{Code availability}
We make our code freely available at the following GitHub repository: \url{https://github.com/HarryMayne/ICU-patient-subgroups}. The code allows researchers with credentialed access to MIMIC-IV to fully recreate our clustering dataset.

% %%%%%%%%%%%%%%%%%%%%%%%%%%%%%%%%%%%%%%%%%%%%%%%%%%%%%%%%%%%%%%%%%
\section{Results}

\subsection{Study population}
The preprocessed data contained 72,896 unique ICU stays and had no missing data. A random sample of 5,000 ICU stays was drawn to derive the clustering solutions. The characteristics of both the random sample and the complete MIMIC-IV cohort are displayed in Table \ref{tab:average_patient_characteristics}. Code status is the only feature where the difference between the mean in the random sample and the remaining data is statistically significant at the 5\% level.

\begin{table}[t!]
\centering
\begin{tabular}{@{}l c c}
\toprule
\multirow{2}{*}{Feature}& \multirow{2}{*}{Random sample} & \multirow{2}{*}{Complete MIMIC-IV cohort}\\
&&\\
\midrule
\multirow{1}{*}{Patient details} & & \\
\quad Age (years) & 64.89 & 64.79 \\
\quad Comorbidity (Charlson Comorbidity Index) & 4.91 & 4.92 \\
\multirow{4}{*}{Hospital admission} & & \\[5ex]
\quad Emergency department admission (\%) & 66.64 & 66.14 \\
\quad Need for surgery (\%) & 43.24 & 42.39 \\
\quad Code status (do not resuscitate) (\%) & 7.08 & 6.33 \\
\quad Severity of illness (LAPS II) & 89.03 & 88.67 \\
\quad Predicted hospital mortality (mean \%) & 13.75 & 13.77 \\[0.5ex]
\multirow{4}{*}{ICU} & & \\[5ex]
\quad Severity of illness (SAPS II) & 35.26 & 35.13 \\
\quad Days of benzodiazepines & 0.68 & 0.63 \\
\quad Days of non-benzodiazepines/other sedatives & 1.11 & 1.06 \\
\quad Days of opiates & 1.70 & 1.65 \\
\multirow{4}{*}{Hospital discharge} & & \\[5ex]
\quad Total length of stay (days) & 11.31 & 10.98 \\
\quad Hospital mortality (\%) & 11.50 & 11.42 \\
\quad Discharged home (\%) & 48.76 & 49.33 \\
\quad Discharged hospice (\%) & 2.12 & 2.18 \\
\quad Discharged skilled facility (\%) & 36.60 & 36.11 \\[0.5ex]
\multirow{4}{*}{Post-discharge} & & \\[5ex]
\quad Death within 30 days of admission  (\%) & 13.82 & 13.83 \\
\quad Readmission within 30 days (\%) & 19.26 & 18.94 \\
\bottomrule
\end{tabular}
\vspace{5mm}
\caption[]{\color{black}{\bf Patient characteristics in the random sample and the complete MIMIC-IV cohort.} The random sample contains 5,000 ICU stays. Except for code status, the means of all features in the random samples are found to be insignificantly different from the remaining data at the 5\% level. Here LAPS~II stands for Laboratory Acute Physiology Score II; and SAPS~II for Simplified Acute Physiology Score II.} 
\label{tab:average_patient_characteristics}
\end{table}

\subsection{Selecting the most appropriate clustering solution}
\begin{figure}[t!]
    \centering
    \includegraphics[width=\textwidth]{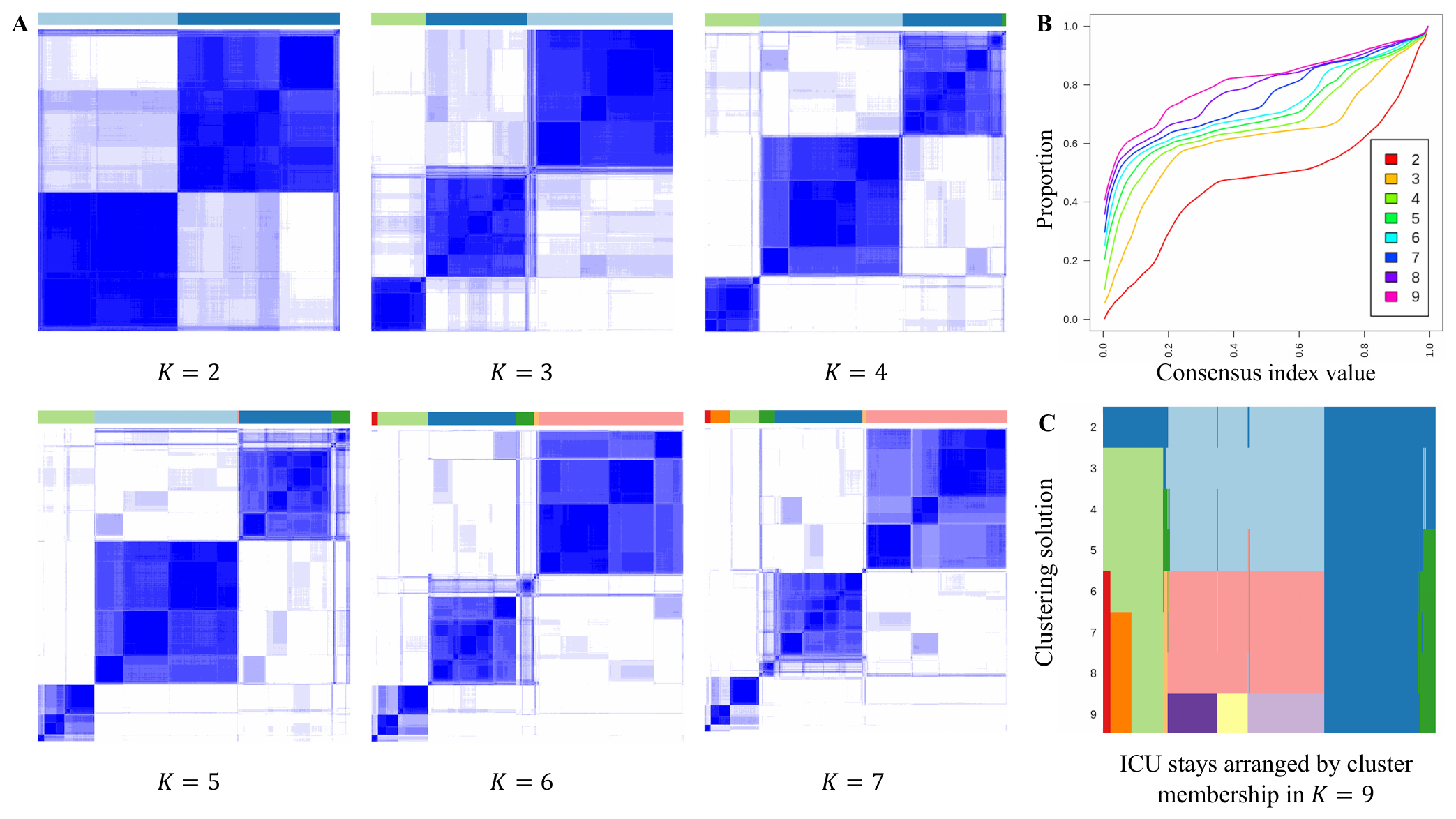}
    \vspace{-5mm}
    \caption[]{\color{black}{\bf Consensus clustering results.} {\bf  A:} The ordered consensus matrices show the stability of the clustering solutions from $K=2$ to $K=7$. Darker shading shows that a pair of examples were more frequently clustered together across the iterations. Therefore, cleaner, sharper matrices represent more stable clustering solutions. The colour bars above the plots show the partitioning of the data into clusters. The $K=8$ and $K=9$ matrices are significantly less clean and not shown. 
    {\bf B:} The CDFs show the proportion of ICU stays with unstable cluster memberships. Each CDF plots the cumulative distribution of indices in the corresponding ordered consensus matrix. A more stable clustering solution would have a higher proportion of consensus indices near 0 and 1 (white and dark blue on the ordered consensus matrices). This corresponds to a more step-like CDF. 
    {\bf C:} The tracking plot shows how cluster membership changes as $K$ increases from $2$ to $9$ (read from the top downwards). Unstable partitions may be visible at one level of the plot and then disappear as the granularity of clustering increases. For instance, a cluster at one level might be amalgamated into a larger cluster at a more granular level. Further details about interpreting these plots can be found in \cite{monti_consensus_2003, senbabaoglu_critical_2014, wilkerson_consensusclusterplus_2010}.
    }
    \label{fig:consensus_clustering_results_plot}
\end{figure}

The consensus clustering results are shown in Figure~\ref{fig:consensus_clustering_results_plot}. When jointly considering the ordered consensus matrices, the CDFs of the consensus indices and the tracking plot, the results suggest that $K=3$ is the most appropriate clustering solution. Considering each visualisation individually, the ordered consensus matrices suggest that $K=3$ or $K=4$ may be the most appropriate solution (Figure~\ref{fig:consensus_clustering_results_plot}A). However, when $K=4$, the smallest cluster is significantly smaller than the other three, a known indicator of an unstable cluster when employing consensus clustering with an agglomerative hierarchical inner loop and average linkage \cite{senbabaoglu_critical_2014}. A qualitative assessment of the CDFs suggests that $K=3$ has the most step-like curve (Figure~\ref{fig:consensus_clustering_results_plot}B). Similarly, the tracking plot also suggests $K=3$ is most appropriate (Figure~\ref{fig:consensus_clustering_results_plot}C). By increasing the granularity of clustering from $K=3$, the clusters split into smaller clusters whilst maintaining the boundaries of the $K=3$ clusters. Therefore, the $K=3$ partitioning can be seen at all subsequent clustering levels. This is strong evidence that the clustering solution is a meaningful division of ICU stays. In contrast, the partitioning where $K=4$ is not consistently found at more granular levels because the smallest cluster is amalgamated into a larger cluster when $K=5$, suggesting the $K=4$ solution is unstable.

\subsection{Cluster characteristics}\label{Sec:our_clusters}
The characteristics of the clusters in the $K=3$ solution are displayed in Table \ref{tab:cluster_characteristics_original}. Cluster 1 ($48.18\%$) contains younger patients (59.20 years) who present with low comorbidity (3.89 CCI) and less severe illnesses (76.30 LAPS II). They have the shortest hospital length of stay (LOS) (7.83 days), receive the least sedation in ICU, and are almost entirely discharged home ($98.42\%$). Cluster 2 ($33.68\%$) contains significantly older patients (68.23 years) who present with higher levels of comorbidity (5.32 CCI) and more severe illnesses (92.23 LAPS II). They have the highest surgery rate (52.97\%), the longest hospital LOS (16.97 days), and require high levels of sedation. The majority of patients survive (97.15\%) but are almost all discharged to skilled facilities (96.97\%) and have high rates of readmission (25.48\%). Cluster 3 ($18.14\%$) contains older patients (73.81 years) in poor prior health (6.90 CCI) and experiencing catastrophic illnesses (116.91 LAPS II). They have the lowest surgery rate (20.95\%), likely an indication of low physiological reserve, and the majority die within 30 days of hospital admission ($76.19\%$).

\begin{table}[t!]
\centering
\begin{tabular}{@{}lccc}
\toprule
& \multicolumn{3}{c}{Clusters} \\
\cline{2-4}
\addlinespace[3pt]
&1&2&3\\
& 48.18\% & 33.68\% & 18.14\%  \\
\midrule
\multirow{1}{*}{Patient details} & \textcolor{white}{clusters1} &\textcolor{white}{clusters1}  &\textcolor{white}{clusters1}  \\
\quad Age (years) & \cellcolor{green!25}59.20 & \cellcolor{orange!25}68.23 & \cellcolor{red!65}73.81 \\
\quad Comorbidity (Charlson Comorbidity Index) & \cellcolor{green!25}3.89 & \cellcolor{orange!25}5.32 & \cellcolor{red!65}6.90\\[0.5ex]
\multirow{4}{*}{Hospital admission} &  &  &   \\[5ex]
\quad Emergency department admission (\%) & \cellcolor{green!25}61.39 & \cellcolor{orange!25}66.86 & \cellcolor{red!65}80.15  \\
\quad Need for surgery (\%) & \cellcolor{orange!25}44.83 & \cellcolor{red!65}52.97 & \cellcolor{green!25}20.95 \\
\quad Code status (do not resuscitate) (\%) & \cellcolor{green!25}0.00 & \cellcolor{orange!25}0.12 & \cellcolor{red!65}38.81  \\
\quad Severity of illness (LAPS II) & \cellcolor{green!25}76.30 & \cellcolor{orange!25}92.23 & \cellcolor{red!65}116.91  \\
\quad Predicted hospital mortality (mean \%) & \cellcolor{green!25}8.23 & \cellcolor{orange!25}14.92 & \cellcolor{red!65}26.23 \\
\multirow{4}{*}{ICU} &  &  &  \\[5ex]
\quad Severity of illness (SAPS II) & \cellcolor{green!25}29.70 & \cellcolor{orange!25}36.94 & \cellcolor{red!65}46.89\\
\quad Days of benzodiazepines & \cellcolor{green!25}0.36 & \cellcolor{orange!25}0.98 & \cellcolor{red!65}1.00\\
\quad Days of non-benzodiazepines/other sedatives & \cellcolor{green!25}0.60 & \cellcolor{red!65}1.75 & \cellcolor{orange!25}1.25  \\
\quad Days of opiates & \cellcolor{green!25}1.00 & \cellcolor{red!65}2.36 & \cellcolor{orange!25}2.34\\
\multirow{4}{*}{Hospital discharge} &  &  &   \\[5ex]
\quad Total length of stay (days) & \cellcolor{green!25}7.83 & \cellcolor{red!65}16.97 & \cellcolor{orange!25}10.04 \\
\quad Hospital mortality (\%) & \cellcolor{green!25}0.00 & \cellcolor{orange!25}2.85 & \cellcolor{red!65}58.10 \\
\quad Discharged home (\%)& \cellcolor{green!25}98.42 & \cellcolor{red!65}0.12 & \cellcolor{orange!25}7.17   \\
\quad Discharged skilled facility (\%)& \cellcolor{green!25}0.00 & \cellcolor{red!65}96.97 & \cellcolor{orange!25}22.81\\
\quad Discharged hospice (\%) & \cellcolor{green!25}0.00 & \cellcolor{green!25}0.00 & \cellcolor{red!65}12.38\\
\multirow{4}{*}{Post-discharge} &  &  & \\[5ex]
\quad 
Death within 30 days of admission  (\%) & \cellcolor{green!25}0.00 & \cellcolor{green!25}0.00 & \cellcolor{red!65}76.19\\
\quad Readmission within 30 days (\%) & \cellcolor{orange!25}19.34 & \cellcolor{red!65}25.48 & \cellcolor{green!25}7.50  \\
\bottomrule
\end{tabular}
\vspace{5mm}
\caption[]{\color{black}{\bf Feature characteristics of the clusters in the $\mathbf{\textit{K}=3}$ solution.} Values highlighted in red represent the cluster with the most medically severe value for each feature. Values highlighted in green represent the least medically severe value. ANOVA and $\chi^2$ tests confirm that all features are non-uniformly distributed  across the clusters at the 1\% significance level, confirming that the clusters have distinct ICU trajectories. A small number of patients ($45$) had missing discharge locations despite surviving their stays, which explains why the discharge location features and hospital mortality do not sum to $100\%$ (see Section \ref{App:feature_matching} of the Appendix for further details). Here LAPS~II stands for Laboratory Acute Physiology Score II; and SAPS~II for Simplified Acute Physiology Score II.} \label{tab:cluster_characteristics_original}
\end{table}

\section{Discussion}

\subsection{Main finding}
In terms of both the number and nature of the clusters identified, the MIMIC-IV clustering results show limited similarities with the study by Vranas et al. \cite{vranas_identifying_2017}, which used data from Kaiser Permanente Northern California hospitals. Since we closely replicated their methodology, we isolated the dataset as the only source of variation and set up a precise test of generalisability. Our results offer evidence against the hypothesis that common patient subgroups exist across different ICU cohorts.

\subsection{Clustering similarity}

Our investigation into the MIMIC-IV ICU cohort identified three distinct clusters (Section~\ref{Sec:our_clusters}), contrasting with the six clusters derived by Vranas et al. \cite{vranas_identifying_2017}. In their analysis, outlined in Table \ref{tab:vranas_results} (Section \ref{App:Vranas} of the Appendix), the authors identified varying patient profiles. These included clusters representing relatively healthy, short-stay ICU patients (Cluster 1); older individuals suffering catastrophic illnesses (Cluster 2); postsurgical and postprocedural patients (Cluster 3); older patients requiring long-term care upon discharge (Cluster 4); previously healthy patients experiencing prolonged stays but ultimately showing good recovery (Cluster 5); and patients with severe illness who expressed a preference for limitations on life-sustaining therapy (Cluster 6).

The primary difference between the two studies is the number of clusters identified. There is no evidence that the MIMIC-IV ICU data contains more than three meaningful clusters. We explored the clustering solution where $K=6$, but found it to be unstable, indicating that the subgroups identified were unlikely to be meaningful. This result also highlights why forcing our clustering to have the same number of subgroups as \cite{vranas_identifying_2017} ex-ante could have led to misleading clustering results, since the derived clusters would not have been representative of subgroups intrinsic to the data.

Next, under the supervision of a highly experienced ICU clinician, we compared the mean characteristics of each of our clusters with those found in \cite{vranas_identifying_2017} to try to identify whether there were any pairs of clusters with similar characteristics. This could have shown that portions of the clustering solutions were similar, even if the overall clustering solutions were different. However, we were unable to find any evidence of similar subgroups. 

We also explored the possibility of there being a many-to-one mapping between clusters. Hypothetically, the two clustering solutions could have identified similar structures in the patient cohorts, albeit at different levels of granularity. This would be consistent with the observations of both a different number of clusters and no pairings of similar clusters.
To examine this possibility, we attempted to map the clusters in \cite{vranas_identifying_2017} to ours in a many-to-one fashion; however, we could not find any evidence of such a mapping. This suggests that the clusters identified in this study represent distinct subgroups to those found in \cite{vranas_identifying_2017} and provides evidence against clustering generalisation.

\subsection{Implications}
Our results have important implications for how ICU restructuring might be operationalised. Since our methodology controlled for differences in clustering approaches and therefore isolated the data as the only source of variation, our results suggest that different ICUs can have significant differences between them, agreeing with prior comparative findings \cite{seymour2012hospital, knaus1993variations} and the expectations of previous ICU patient clustering studies \cite{de2024deep}. As a consequence, whilst a clustering solution might be an accurate representation of patient subgroups in one ICU, it may not represent the best subgroups in another. The theoretical efficiency benefits of ICU restructuring depend on the subunits matching the subgroups present in the ICU patients, since this creates groups with lower heterogeneity.  Therefore, the potential efficiency gains might be greater if the number and nature of the subunits were tailored to each ICU individually. Conversely, a standardised restructuring approach, where each ICU is restructured into a common set of subunits, may try to group patients into subgroups which do not exist and do a poorer job of reducing heterogeneity.

These findings also relate to alternative approaches to test for clustering generalisability. In particular, de Kok et al. \cite{de2024deep} explored whether clusters derived in one ICU dataset might extrapolate to a different dataset, finding positive generalisability results. If we had taken an equivalent approach of training a clustering model on one dataset and applying the trained model to a new dataset, we would have identified six clusters which may or may not have been similar to those in \cite{vranas_identifying_2017}. However, our results for $K=6$ showed high levels of instability, suggesting that such an approach would have derived clusters which were unrepresentative of stable and meaningful subgroups intrinsic to the MIMIC-IV patients. If the aim of ICU clustering is to reduce patient heterogeneity by uncovering meaningful patient subgroups, then the approach taken by de Kok et al. \cite{de2024deep} may be practical, but is unlikely to be optimal, since it does not fully account for differences between ICU populations.

Our results also make more general contributions to ICU patient clustering, since they are further evidence that it is possible to use unsupervised learning to identify meaningful subgroups of ICU patients. This supports previous studies which have derived patient subgroups to either assess the heterogeneity of ICU patients \cite{hyun_exploration_2020, merkelbach_novel_2023, shi_identifying_2023}, work towards personalised medicine \cite{castela_forte_identifying_2021, merkelbach_novel_2023, fuest2023clustering}, or more explicitly advance ICU restructuring \cite{castela_forte_identifying_2021, vranas_identifying_2017, de2024deep}.

\subsection{Limitations and future research directions}\label{sec:limitations_future_research}
Our study has several limitations. First, whilst we made efforts to replicate the features in Vranas et al. \cite{vranas_identifying_2017}, we could not achieve exact replication due to data constraints and incomplete information about their methodology. This means that there are minor discrepancies in feature definitions (see Section \ref{App:feature_matching} of the Appendix). 
We anticipate that the impact of this should be limited, since analogous features should represent the same underlying concepts and the consensus clustering method used in our study is inherently robust to minor variations in the features \cite{monti_consensus_2003}. 

Second, there is some variation between the MIMIC-IV patients and the cohort in \cite{vranas_identifying_2017}. On average, the MIMIC-IV patients are slightly more unwell, since they receive higher levels of treatment, have longer ICU stays and a higher mortality rate. A full comparison is shown in Table \ref{tab:vranas_characteristics_vs_Mimic} in the Appendix. Whilst excessive variation would be a concern, some degree of variation in ICU populations is expected and necessary for a realistic test of generalisability.

Third, clustering only attempts to identify the best way to group patients, which does not rule out the existence of many other good partitions to reduce patient heterogeneity \cite{de2024deep}. Our results indicate that the MIMIC-IV cohort does not intrinsically have similar subgroups to those in \cite{vranas_identifying_2017}, which means that partitioning the MIMIC-IV patients into the clusters in \cite{vranas_identifying_2017} would not be the optimal way to reduce patient heterogeneity. However, this does not mean that such a division is not a good or useful way to reduce heterogeneity. This implies that standardised ICU restructuring could still lead to potential efficiency gains, albeit fewer than in a tailored restructuring approach. If the difference in efficiency gains is small, then a standardised approach may still be a good solution. This would be especially relevant if the process of tailoring the restructuring to each individual ICU was associated with significant administrative costs and inefficiencies. However, our results and this discussion motivate further questions about the standardised ICU restructuring approach. For instance, since we identified different clusters to Vranas et al. \cite{vranas_identifying_2017}, which set of clustering results should a standardised restructuring approach use?
A key direction for future research is to design a framework to quantify the potential efficiency gains from restructuring in different scenarios.

Fourth, whilst not the focus of our study, there are existing criticisms of this restructuring approach which have yet to be addressed in the literature. Notably, Kramer \cite{kramer_group_2017} argued that, since the features span the duration of patients' ICU stays, patients could only be triaged to subunits retrospectively. This is an important critique and highlights how the potential efficiency gains from any ICU restructuring may be bottlenecked by the ability of clinicians to accurately triage patients to the subunits at ICU admission. In this paper, we proposed that identifying comprehensive patient subgroups first, then designing triage methods later, has the potential to ultimately lead to better subgroups than if subgroups were based solely on features available at ICU admission. This is because resource reallocation would be optimised to match patients' medical need in ICU rather than at an earlier point. There are many possible directions to explore how machine learning might support clinicians to triage patients. For example, a supervised model could be trained to predict subunit allocation using only data collected before ICU admission. Since pathologies often precede ICU admission, this data may be sufficiently rich to accurately predict subunit allocation. Alternatively, it may be possible to recreate our clusters using features collected before ICU admission. Patient triage is an important avenue for future research to address.

\section{Conclusion}

In this paper, we tested whether the clusters identified in one ICU population could be identified in a different population. We explored whether the clustering solution in Vranas et al. \cite{vranas_identifying_2017} would generalise to the MIMIC-IV dataset. Our results suggest that there is limited similarity between the two sets of results, providing evidence against the hypothesis of generalisability. These findings suggest that a standardised approach to ICU clustering is unlikely to be appropriate because there is too much variation between ICU populations. The potential efficiency gains from ICU restructuring might be greater if the number and nature of the subunits were tailored to each ICU individually. Future research should attempt to quantify the potential benefits of these proposals.

\section*{Acknowledgements}
This project has benefited from support to HM by the Economic and Social Research Council grant for Digital Social Science [ES/P000649/1]. We thank Felix Krones, Yushi Yang, and Karolina Korgul for their helpful discussions and feedback on this work. 

\bibliography{refs}
\bibliographystyle{unsrtnat}

\newpage 
\appendix

\section{Feature matching}\label{App:feature_matching}

To ensure that our method was a strong test of the generalisability of Vranas et al. \cite{vranas_identifying_2017}, we took care to recreate their clustering features as closely as possible. The majority of features were relatively straightforward to recreate but six required significant value judgement: comorbidity, severity of illness at hospital admission (LAPS II), predicted hospital mortality at hospital admission (based on LAPS II), code status, severity of illness at ICU admission (SAPS II) and discharge location. In this section we describe how each feature was recreated, focusing more on those that were challenging. A comparison between the two sets of features is discussed and shown in Table \ref{tab:vranas_characteristics_vs_Mimic}.

\textbf{Age:} We build age using the MIMIC-IV code repository of common features \cite{johnson_mimic_2018}. Approximately $3\%$ of the patients in MIMIC-IV were older than 89 and had their ages coded as 91 to prevent re-identification. These patients were left in our dataset without adjustment. 

\textbf{Comorbidity:} Vranas et al. \cite{vranas_identifying_2017} used Comorbidity Point Score, version 2 (COPS II) to measure the levels of patient comorbidity. This score was created by researchers at the Kaiser Permanente Northern California \cite{escobar_risk-adjusting_2013} and proved especially challenging to attempt to recreate. Significant difficulties meant that we were not able to recreate this feature. For example, COPS II relies on patient diagnoses at admission, however, MIMIC-IV only contains diagnoses at hospital discharge. Although we experimented with using a version of discharge diagnosis as a proxy, creating a suitable mapping proved infeasible, since patients would have many diagnoses at hospital discharge listed. This would have added significant noise because we would have had to first determine the primary diagnosis and second assume that this was also the primary diagnosis at admission.

\begin{figure}[t!]
\hspace*{1cm}
\begin{center}
\begin{adjustbox}{width=0.6\textwidth, trim=0\width{} 0pt 0pt 0pt,clip}
\begin{tikzpicture}
\node[anchor=south west,inner sep=0] (image) at (0,0) {\includegraphics[width=\textwidth]{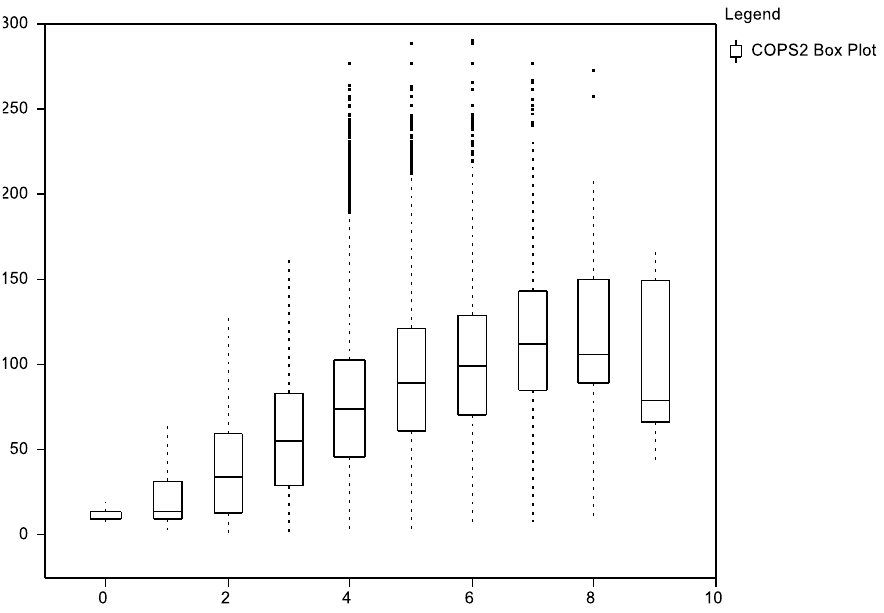}};
\begin{scope}[x={(image.south east)},y={(image.north west)}]
\node[rotate=90, font=\Large] at (-0.05, 0.5) {COPS II score};
\node[font=\Large] at (0.45, -0.05) {Charlson Comorbidity Index (CCI)};
\end{scope}
\end{tikzpicture}
\end{adjustbox}
\end{center}
\caption[]{\color{black}{\bf The relationship between COPS II and CCI.} It appears that CCI is a relatively good substitute for COPS II because they have an approximately linear relationship. A limitation is that this trend breaks down for higher CCI scores. Despite differences, both should capture similar underlying dynamics. This figure is taken from the supplementary methods of Escobar et al. \cite{escobar_risk-adjusting_2013}.}
\label{fig:CCI_vs_COPSII}
\end{figure}

Since recreating COPS II was ultimately not feasible, we used the Charlson Comorbidity Index (CCI) as a substitute. Specifically, we used the Deyo implementation of CCI \cite{deyo_adapting_1992}, an adapted version of the original index \cite{charlson_new_1987}. We chose CCI for two reasons: first, it is present in the MIMIC-IV code repository of common features \cite{johnson_mimic_2018}, second, the original COPS II paper explored the relationship between COPS II and CCI, and showed that they are relatively closely related (Figure \ref{fig:CCI_vs_COPSII}).

\textbf{Emergency department admission:} It was challenging to recreate Vranas et al.'s \cite{vranas_identifying_2017} implementation of emergency department admission, since it was unclear how it was defined. We considered multiple operationalisations and chose the one which had the most similar rate of occurrence as Vranas et al. \cite{vranas_identifying_2017}. We note that this is not a perfect heuristic, but our other candidate features had significantly different rates of occurrence, hence it was a natural choice. Our feature is defined as all patients with recorded emergency department admission and discharge times. Patients where the emergency department discharge time was recorded as occurring before their admission, and patients who went to the emergency department after their hospital admission, were not included as having emergency department admissions. We wish to highlight that our feature is different from the MIMIC-IV feature `admission location' (found in the admissions table in the \textit{hosp} module), which contains a category for admission via the `emergency room'. Unexpectedly, we found there to be significant dissimilarity between the two candidate features.

\textbf{Need for surgery:} Our interpretation of patients' need for surgery is whether or not their admission to hospital occurred under a surgical service. The following categories of care were included as surgical services: surgical, cardiac surgery, neurological surgical, orthopaedic surgical, thoracic surgical and vascular surgical. This may differ from Vranas et al. \cite{vranas_identifying_2017}, since our feature only considers patients admitted for surgery rather than all patients who received surgery. However, we also found that the rate of occurrence was significantly higher in our data, so this is unlikely to be the case.

\textbf{Code status:} Vranas et al. \cite{vranas_identifying_2017} used three categories for code status: full code, partial code and do not resuscitate (DNR) \cite{vranas_identifying_2017}. However, it was unclear how partial code was defined, and further, there was no analogous code status category in MIMIC-IV. As a result, we coded all patients as either full code or DNR. 

Approximately 55\% of values were missing and were assumed to be full code, since this is the medical default. If patients had multiple code statuses over their hospitalisations, they were coded as DNR.

By excluding partial code, all patients were either full code or DNR. Since both features encoded the same information, only DNR was included in the clustering. This may be a significant deviation from Vranas et al. \cite{vranas_identifying_2017} as it may have reduced the influence of code status in the final clustering.

\textbf{Severity of illness at hospital admission (LAPS II):} Severity of illness at hospital admission was the most challenging feature to recreate. LAPS II was created by researchers at the Kaiser Permanente Northern California \cite{escobar_risk-adjusting_2013, liu_length_2010} and uses 5 vital signs and 18 laboratory test results to produce a final score. Since there was no obvious alternative to this feature in MIMIC-IV, we invested significant time to construct it from more granular data. 

The feature was built in two stages: a preliminary stage to categorise whether patients were low-risk or high-risk, then a secondary stage to build the LAPS II score, using patient risk status to direct data imputation.

\textbf{LAPS II preliminary stage:} The preliminary stage used six common features to build a simple measure of predicted mortality. These were age at admission, gender, emergency department admission, blood urea nitrogen (BUN)/creatinine ratio, sodium and anion gap/serum bicarbonate ratio.

We used a logistic regression model with predefined coefficients to assign each ICU stay a predicted mortality \cite{escobar_risk-adjusting_2013}. Next, a threshold of $0.06$ was implemented to divide low-risk from high-risk patients. This cut-off was chosen to match the original implementation \cite{escobar_risk-adjusting_2013}.

\textbf{LAPS II secondary stage:} The secondary stage used 21 features, some of which overlapped with the preliminary stage, to build the final LAPS II score. These features were all laboratory results collected within the first 72 hours of hospital admission. In cases where multiple results had been collected across those 72 hours, the most medically severe value was chosen. Feature values were then assigned points based on severity and the points were summed to generate a final score. If the patient was low-risk (from the preliminary model), missing data were imputed with `normal' test results. If the patient was high-risk, missing data were imputed with more medically severe test results. Further details can be found in the original paper \cite{escobar_risk-adjusting_2013}.

Our only major deviations from the original method were using troponin T as a substitute for troponin I (adjusting the thresholds appropriately) and using the Glasgow Coma Scale for neurological score.

\textbf{Predicted hospital mortality:} We built predicted hospital mortality using the LAPS II score following the implementation in Lagu et al. \cite{lagu_validation_2016}. A logistic regression model of the following form was used to calibrate LAPS~II
\begin{equation}
\begin{split}
\text{{Mortality}} = \beta_0 &+ \beta_1\text{{age}} + \beta_2\text{{gender}} + \beta_3\text{{race}} + \beta_4\text{{LAPSII}} + \beta_5\text{{LAPSII}}^2,
\end{split}
\end{equation}
where $\text{{Mortality}}$ is mortality 30 days after hospital admission.

\textbf{Severity of illness at ICU admission (SAPS II):} Vranas et al. \cite{vranas_identifying_2017} used eSAPSIII, an electronic adaptation of the Simplified Acute Physiology Score, version 3. We approximated this with SAPS II \cite{le1993new}, since the code implementation was provided in the MIMIC-IV code repository \cite{johnson_mimic_2018}. Known differences between SAPS II and SAPS III are a limitation of this approach \cite{mit_critical_data_mortality_2016}.

\textbf{Days of benzodiazepines:} To build this feature we considered the administration of Diazepam (Valium), Lorazepam (Ativan) or Midazolam (Versed). We defined the feature as the discrete number of days on which a patient was given benzodiazepines \cite{vranas_identifying_2017}.

\textbf{Days of non-benzodiazepines/other sedatives:} This feature included all non-benzodiazepines and other sedatives, excluding benzodiazepines and opiates. We included Propofol (intubation), Propofol, Dexmedetomidine (Precedex), Pentobarbital, Ketamine, Ketamine (intubation), Haloperidol (Haldol), Nitrous Oxide (inhaled), Sevoflurane (inhaled), Isoflurane (inhaled), Etomidate (intubation) and Phenobarbital.

\textbf{Days of opiates:} We included Fentanyl (concentrate), Morphine Sulfate, Meperidine (Demerol), Fentanyl (push), Hydromorphone (Dilaudid), Methadone Hydrochloride and Fentanyl.

\textbf{Discharge location:} Vranas et al. \cite{vranas_identifying_2017} grouped discharge location into three categories: home, skilled nursing facility and hospice. The MIMIC-IV data contains nine categories; thus, we mapped these to the three categories mentioned. Our mapping is as follows: home (home health care, home, against advice, assisted living), skilled nursing facility (skilled nursing facility, rehab, chronic/long term acute care, psych facility, acute hospital, other facility, healthcare facility), hospice (hospice). A small number of patients (less than $1\%$) had missing discharge locations and were not recorded to have died in hospital. This explains why the discharge and hospital mortality features in the tables do not sum to 1. From analysing these patients in more detail, it appears that they are far healthier than other patients, presenting with lower ages, lower LAPS II, and far lower levels of ICU treatment. The majority of these patients were clustered into the first cluster. 

\textbf{Other features:} Hospital length of stay, mortality, mortality within 30 days and readmission within 30 days were defined in the expected manner.

\textbf{Comparison:} The mean values for features in Vranas et al. \cite{vranas_identifying_2017} and our dataset are compared in Table \ref{tab:vranas_characteristics_vs_Mimic}. In general, the patients in the MIMIC-IV cohort are more unwell. They have a higher severity of illness upon admission and a higher predicted hospital mortality. Patients in MIMIC-IV consistently receive more treatments in ICU. Mortality rates are higher in MIMIC-IV and a higher proportion of patients are discharged to skilled facilities.

There are some notable discrepancies which are likely caused by different feature definitions. The need for surgery in MIMIC-IV is 42.4\%, compared to 24.8\% in Vranas et al. \cite{vranas_identifying_2017}. LAPS II and predicted mortality differ, which is likely a result of the complex recreation process. The percentage of patients discharged to skilled facilities in MIMIC-IV is 36.1\%, compared to 15.6\% in Vranas et al. \cite{vranas_identifying_2017}. This could be a result of Vranas et al. \cite{vranas_identifying_2017} categorising low-severity discharge locations as `home' rather than `skilled facility'.

\begin{table}[t!]
\centering
\begin{tabular}{@{}l c c}
\toprule
\multirow{2}{*}{Feature}&\multirow{2}{*}{Vranas et al. \cite{vranas_identifying_2017}}&\multirow{2}{*}{MIMIC-IV cohort}\\
&&\\
\midrule
\multirow{1}{*}{Patient details} & & \\
\quad Age (years) & 65.2 & 64.8 \\
\quad Comorbidity & - & - \\
\multirow{4}{*}{Hospital admission} & & \\[5ex]
\quad Emergency department admission (\%) & 74.9 & 66.1 \\
\quad Need for surgery (\%) & 24.8 & 42.4 \\
\quad Code status (do not resuscitate) (\%) & 9.3$^{*}$ & 6.3 \\
\quad Severity of illness (LAPS II) & 82.4 & 88.7 \\
\quad Predicted hospital mortality (mean \%) & 6.8 & 13.8 \\[0.5ex]
\multirow{4}{*}{ICU} & & \\[5ex]
\quad Severity of illness & - & - \\
\quad Days of benzodiazepines & 0.2 & 0.6 \\
\quad Days of non-benzodiazepines/other sedatives & 0.5 & 1.1 \\
\quad Days of opiates & 0.3 & 1.7 \\
\multirow{4}{*}{Hospital discharge} & & \\[5ex]
\quad Total length of stay (days) & 8.3 & 11.0 \\
\quad Hospital mortality (\%) & 10.2 & 11.4 \\
\quad Discharged home (\%) & 71.2 & 49.3 \\
\quad Discharged hospice (\%) & 2.0 & 2.2 \\
\quad Discharged skilled facility (\%) & 15.6 & 36.1 \\[0.5ex]
\multirow{4}{*}{Post-discharge} & & \\[5ex]
\quad Death within 30 days of admission  (\%) & 12.3 & 13.8 \\
\quad Readmission within 30 days (\%) & 18.7 & 19.0 \\
\bottomrule
\end{tabular}
\vspace{5mm}
\caption[]{\color{black}{\bf Patient characteristics in Vranas et al. \cite{vranas_identifying_2017} and the MIMIC-IV cohort.} The characteristics of patients in Vranas et al.'s \cite{vranas_identifying_2017} data differ from the MIMIC-IV cohort. In general, patients in the MIMIC-IV cohort are more unwell, receive higher levels of treatment and are more likely to die. Redacted features are those where a different definition was used between studies. Here LAPS~II stands for Laboratory Acute Physiology Score II. $^{*}$This proportion includes patients defined as partial code in Vranas et al. \cite{vranas_identifying_2017}.} 
\label{tab:vranas_characteristics_vs_Mimic}
\end{table}

\section{Data preprocessing}\label{App:preprocessing}
The data were passed through a series of filters to detect and manage outliers. Medically illogical values were removed and recorded as missing. In some cases, the data error was obviously attributable to a recording error and the correct value could be imputed. For instance, some data that should have been recorded as a decimal was instead recorded as a percentage, or, some data were very clearly recorded with the wrong units. In these cases, the incorrect values were replaced by the implied correct value.

Furthermore, the data were checked for clear contradictions and appropriately corrected. The criteria for the checks are detailed in Table \ref{tab:sense_checks}.

\begin{table}[t!]
\centering
\begin{tabular}{lll}
\toprule
\multirow{2}{*}{Check criteria} & \multirow{2}{*}{Amendment to the dataset} & \multirow{2}{*}{ICU stays affected} \\
&&  \\
\midrule
Total LOS less than ICU LOS & Total LOS set to ICU LOS & 4,538 \\
Mortality in hospital and patient discharged & Discharge location set as null & $87$ \\
Mortality in hospital and readmission & Readmission set as null & $108$ \\
\bottomrule
\end{tabular}
\vspace{5mm}
\caption[]{\color{black}{\bf Checks for contradictions in the data.} This table shows a series of checks for clear contradictions in the data, the resulting amendments made, and the number of ICU stays that were affected by the changes. In cases where the patient died in hospital and was recorded as being discharged/readmitted, it was assumed that the death record was more likely to be correct. LOS: Length of stay.}
\label{tab:sense_checks}
\end{table}

\section{Clustering results from Vranas et al. [7]} \label{App:Vranas}

The equivalent clustering results from Vranas et al. \cite{vranas_identifying_2017} are shown in Table \ref{tab:vranas_results}. They found six clusters with limited similarities to our results. Their data is not publicly available; thus, we were not able to replicate their findings.

\begin{table}[t!]
\centering
\fontsize{10.5pt}{13.5pt}\selectfont
\begin{tabular}{@{}lcccccc}
\toprule
& \multicolumn{6}{c}{Clusters}\\
\cline{2-7}
\addlinespace[3pt]
&1&2&3&4&5&6\\
&38.7\%&12.4\%&25.0\%&17.9\%&4.1\%&1.8\%\\
\midrule
\multirow{1}{*}{Patient details} &  &  &  & &  &  \\
\quad Age (years) & \cellcolor{greenishyellow!45}60.9 & \cellcolor{yellow!40}72.7 & \cellcolor{lightorange!40}63.8 & \cellcolor{orangered!55}74.8 & \cellcolor{green!25}58.7 & \cellcolor{red!65}79.4 \\
\quad Comorbidity (COPS II) & \cellcolor{greenishyellow!45}44 & \cellcolor{orangered!55}65 & \cellcolor{green!25}35 & \cellcolor{yellow!40}63 & \cellcolor{lightorange!40}48 & \cellcolor{red!65}70 \\[0.5ex]
\multirow{4}{*}{Hospital admission} &  &  &  &  &  &   \\[5ex]
\quad Emergency department admission (\%) & \cellcolor{red!65}100.0 & \cellcolor{lightorange!40}86.8 & \cellcolor{green!25}21.5 & \cellcolor{yellow!40}82.8 & \cellcolor{greenishyellow!45}79.7 & \cellcolor{red!65}100.0 \\
\quad Need for surgery (\%) & \cellcolor{green!25}0.2 & \cellcolor{yellow!40}9.7 & \cellcolor{red!65}76.9 & \cellcolor{lightorange!40}17.2 & \cellcolor{orangered!55}19.8 & \cellcolor{greenishyellow!45}4.4 \\
\quad Code status (do not resuscitate) (\%) & \cellcolor{green!25}0.0 & \cellcolor{orangered!55}18.0 & \cellcolor{green!25}0.0 & \cellcolor{red!65}28.2 & \cellcolor{green!25}0.0 & \cellcolor{green!25}0.0 \\
\quad Severity of illness (LAPS II) & \cellcolor{greenishyellow!45}90 & \cellcolor{orangered!55}120 & \cellcolor{green!25}33 & \cellcolor{lightorange!40}95 & \cellcolor{yellow!40}92 & \cellcolor{red!65}128 \\
\quad Predicted hospital mortality (mean \%) & \cellcolor{greenishyellow!45}4.8 & \cellcolor{orangered!55}16.5 & \cellcolor{green!25}1.9 & \cellcolor{lightorange!40}9.4 & \cellcolor{yellow!40}8.1 & \cellcolor{red!65}22.5 \\
\multirow{4}{*}{ICU} &  &  &  &  &  &  \\[5ex]
\quad Severity of illness (SAPS II) & \cellcolor{greenishyellow!45}8.0 & \cellcolor{red!65}21.6 & \cellcolor{green!25}3.5 & \cellcolor{yellow!40}12.5 & \cellcolor{lightorange!40}13.1 & \cellcolor{orangered!55}16.4 \\
\quad Days of benzodiazepines & \cellcolor{green!25}0.0 & \cellcolor{orangered!55}0.3 & \cellcolor{green!25}0.0 & \cellcolor{yellow!40}0.1 & \cellcolor{red!65}2.1 & \cellcolor{yellow!40}0.1 \\
\quad Days of non-benzodiazepines$^*$ & \cellcolor{green!25}0.2 & \cellcolor{orangered!55}0.8 & \cellcolor{greenishyellow!45}0.3 & \cellcolor{yellow!40}0.4 & \cellcolor{red!65}3.7 & \cellcolor{lightorange!40}0.6 \\
\quad Days of opiates & \cellcolor{green!25}0.1 & \cellcolor{orangered!55}0.7 & \cellcolor{greenishyellow!45}0.2 & \cellcolor{greenishyellow!45}0.2 & \cellcolor{red!65}3.7 & \cellcolor{lightorange!40}0.3 \\
\multirow{4}{*}{Hospital discharge} &  &  &  &  &  &   \\[5ex]
\quad Total length of stay (days) & \cellcolor{green!25}5.1 & \cellcolor{yellow!40}7.0 & \cellcolor{greenishyellow!45}6.2 & \cellcolor{orangered!55}11.1 & \cellcolor{red!65}32.3 & \cellcolor{lightorange!40}7.7 \\
\quad Hospital mortality (\%) & \cellcolor{green!25}0.0 & \cellcolor{red!65}78.6 & \cellcolor{green!25}0.0 & \cellcolor{green!25}0.0 & \cellcolor{lightorange!40}10.1 & \cellcolor{orangered!55}23.1 \\
\quad Discharged home (\%)& \cellcolor{green!25}100.0 & \cellcolor{red!65}5.6 & \cellcolor{green!25}100.0 & \cellcolor{orangered!55}16.5 & \cellcolor{yellow!40}73.9 & \cellcolor{lightorange!40}46.2 \\
\quad Discharged skilled facility (\%)& \cellcolor{green!25}0.0 & \cellcolor{green!25}0.0 & \cellcolor{green!25}0.0 & \cellcolor{red!65}83.5 & \cellcolor{yellow!40}14.0 & \cellcolor{orangered!55}30.8 \\
\quad Discharged hospice (\%) & \cellcolor{green!25}0.0 & \cellcolor{red!65}15.8 & \cellcolor{green!25}0.0 & \cellcolor{green!25}0.0 & \cellcolor{orangered!55}1.9 & \cellcolor{green!25}0.0 \\
\multirow{4}{*}{Post-discharge} &  &  &  &  &  & \\[5ex]
\quad Death within 30 days of admission  (\%) & \cellcolor{green!25}0.0 & \cellcolor{red!65}92.1 & \cellcolor{greenishyellow!45}0.1 & \cellcolor{lightorange!40}4.5 & \cellcolor{yellow!40}1.0 & \cellcolor{orangered!55}27.5 \\
\quad Readmission within 30 days (\%) & \cellcolor{lightorange!40}21.0& \cellcolor{green!25}1.0 & \cellcolor{greenishyellow!45}15.7 & \cellcolor{red!65}28.2 & \cellcolor{yellow!40}17.4 & \cellcolor{orangered!55}22.0 \\
\bottomrule
\end{tabular}
\vspace{5mm}
\caption[]{\color{black}{\bf Cluster characteristics for the derived clusters in Vranas et al. \cite{vranas_identifying_2017}.} The clusters are highlighted based on medical severity where, for each feature, the least severe cluster is shown in green and the most severe in red.
$^*$Non-benzodiazepines/other sedatives. COPS II: Comorbidity Point Score II. Vranas et al. \cite{vranas_identifying_2017} also included `partial code', however, this was merged with `do not resuscitate' for a better comparison with our results.
} \label{tab:vranas_results}
\end{table}

\end{document}